\documentclass{article}
\usepackage[utf8]{inputenc}
\usepackage{amsmath, amsthm, amsfonts, cuted, graphicx}
\usepackage{lineno}
\usepackage[a4paper, total={6in, 8in}]{geometry}
\usepackage{booktabs}
\usepackage{tabularx}
\usepackage{authblk}

\newcommand{\W}{\boldsymbol{W}}
\newcommand{\Omegab}{\boldsymbol{\Omega}}
\newcommand{\Xb}{\boldsymbol{X}}
\newcommand{\Yb}{\boldsymbol{Y}}

\title{Intercomparison of Machine Learning Methods for Statistical Downscaling: The Case of Daily and Extreme Precipitation}

\author[1]{Thomas Vandal\thanks{vandal.t@husky.neu.edu}}
\author[2]{Evan Kodra\thanks{evan.kodra@risq.io}}
\author[1]{Auroop R Ganguly\thanks{a.ganguly@neu.edu}}

\affil[1]{Sustainability and Data Science Lab \\ Civil Engineering Dept Northeastern University}
\affil[2]{risQ Inc.}

\begin{document}

\maketitle



\abstract{
	Statistical downscaling of global climate models (GCMs) allows researchers to study local climate change effects decades into the future. A wide range of statistical models have been applied to downscaling GCMs but recent advances in machine learning have not been explored. In this paper, we compare four fundamental statistical methods, Bias Correction Spatial Disaggregation (BCSD), Ordinary Least Squares, Elastic-Net, and Support Vector Machine, with three more advanced machine learning methods, Multi-task Sparse  Structure Learning (MSSL), BCSD coupled with MSSL, and Convolutional Neural Networks to downscale daily precipitation in the Northeast United States. Metrics to evaluate of each method's ability to capture daily anomalies, large scale climate shifts, and extremes are analyzed. We find that linear methods, led by BCSD, consistently outperform non-linear approaches. The direct application of state-of-the-art machine learning methods to statistical downscaling does not provide improvements over simpler, longstanding approaches.
}


\section{Introduction}

The sustainability of infrastructure, ecosystems, and public health depends on a predictable and stable climate. Key infrastructure allowing society to function, including power plants and transportation systems, are built to sustain specific levels of climate extremes and perform optimally in it's expected climate. Studies have shown that the changing climate has had, and will continue to have, significant impacts on critical infrastructure~\cite{ganguli2015water,neumann2015climate}. Furthermore, climate change is having dramatic negative effects to ecosystems, from aquatic species to forests ecosystems, caused by increases in greenhouse gases and temperatures~\cite{walther2002ecological,parmesan2006ecological,hansen2013high}. Increases in frequency and duration of heat waves, droughts, and flooding is damaging public health~\cite{haines2006climate,frumkin2008climate}. 

Global Circulation Models (GCMs) are used to understand the effects of the changing climate by simulating known physical processes up to two hundred years into the future. The computational resources required to simulate the global climate on a large scale is enormous, limiting models to coarse spatial and temporal scale projections. While the coarse scale projections are useful in understanding climate change at a global and continental level, regional and local understanding is limited. Most often, the critical systems society depends on exist at the regional and local scale, where projections are most limited. Downscaling techniques are applied to provide climate projections at finer spatial scales, exploiting GCMs to build higher resolution outputs. Statistical and dynamical are the two classes of techniques used for downscaling. The statistical downscaling (SD) approach aims to learn a statistical relationship between coarse scale climate variables (ie. GCMs) and high resolution observations. The other approach, dynamical downscaling, joins the coarse grid GCM projections with known local and regional processes to build Regional Climate Models (RCMs). RCMs are unable to generalize from one region to another as the parameters and physical processes are tuned to specific regions. Though RCMs are useful for hypothesis testing, their lack of generality across regions and extensive computational resources required are strong disadvantages.

\subsection{Statistical Downscaling}

Statistical downscaling methods are further categorized into three approaches, weather generators, weather typing, and transfer functions. Weather generators are typically used for temporal downscaling, rather than spatial downscaling. Weather typing, also known as the method of analogues, searches for a similar historical coarse resolution climate state that closely represents the current state. Though this method has shown reasonable results~\cite{frost2011comparison}, in most cases, it is unable to satisfy the non-stationarity assumption in SD. Lastly, transfer functions, or regression methods, are commonly used for SD by learning functional relationships between historical precipitation and climate variables to project high resolution precipitation. 


A wide variety of regression methods have been applied to SD, ranging from Bias Correction Techniques to Artificial Neural Networks. Traditional methods include Bias Correction Spatial Disaggregation (BCSD)~\cite{wood2002long} and Automated Regression Based Downscaling (ASD)~\cite{hessami2008automated} and are the most widely used. BSCD  assumes that the climate variable being downscaled is well simulated by GCMs, which often is not the case with variables such as precipitation~\cite{schiermeier2010real}. Rather than relying on projections of the climate variable being downscaled, regression methods can be used to estimate the target variable. For instance, precipitation can be projected using a regression model with variables such as temperature, humidity, and sea level pressure over large spatial grids. High dimensionality of covariates leads to multicollinearity and overfitting in statistical models stemming from a range of climate variables over three dimensional space. ASD improves upon multiple linear regression by selecting covariates implicitly, using covariate selection techniques such as backward stepwise regression and partial correlations.  The Least Absolute Shrinkage and Selection Operator (Lasso), a widely used method for high dimensional regression problems through the utilization of a $l_1$ penalty term, is analogous to ASD and has shown superior results in SD~\cite{tibshirani1996regression,hammami2012predictor}. Principle component analysis (PCA) is another popular approach to dimensionality reduction in SD~\cite{tatli2004statistical,Ghosh2010,jakob2011empirical}, decomposing the features into a lower dimensional space to minimize multicollinearity between covariates. PCA is disadvantaged by the inability to infer which covariates are most relevant to the problem, steering many away from the method. Other methods for SD include Support Vector Machines (SVM)~\cite{Ghosh2010}, Artificial Neural Networks (ANNs)~\cite{taylor2000quantile,Coulibaly2005}, and Bayesian Model Averaging~\cite{Zhang2015}.

Many studies have aimed to compare and quantify a subset of the SD models presented above by downscaling averages and/or extremes at a range of temporal scales. For instance, Burger et al. presented an intercomparison on five state-of-the-art methods for downscaling temperature and precipitation at a daily temporal resolution to quantify extreme events~\cite{Burger2012}. Another recent study by Gutmann et al. presented an intercomparison of methods on daily and monthly aggregated precipitation~\cite{gutmann2014intercomparison}. These studies present a basis for comparing SD models by downscaling at a daily temporal resolution to estimate higher level climate statistics, such as extreme precipitation and long-term droughts. In this paper we follow this approach to test the applicability of more advanced machine learning models to downscaling.

\subsection{Multi-task Learning for Statistical Downscaling}

Traditionally, SD has focused on downscaling a locations independently without accounting for clear spatial dependencies in the system. Fortunately, numerous machine learning advances may aid SD in exploiting such dependencies. Many of these advancements focus on an approach known as multi-task learning, aiming to learn multiple tasks simultaneously rather than in isolation. A wide variety of studies have shown that exploiting related tasks through multi-task learning (MTL) greatly outperforms single-task models, from computer vision~\cite{zhang2012robust} to biology~\cite{kim2010tree}.  Consider the work presented by~\cite{evgeniou2007multi} in which increasing the number of tasks leads to more significant feature selection and lower test error through the inclusion of task relatedness and regularization terms in the objective function. MTL has also displayed the ability to uncover and exploit structure between task relationships~\cite{zhang2012convex,chen2011integrating,argyriou2007spectral}. 

Recently Goncalves et al. presented a novel method, Multi-task Sparse Structure Learning (MSSL), ~\cite{goncalves2014multi} and applied it to GCM ensembles in South America. MSSL aims to exploit sparsity in both the set of covariates as well as the structure between tasks, such as set of similar predictands, through alternating optimization of weight and precision (inverse covariance) matrices. The results showed significant improvements in test error over Linear Regression and Multi-model Regression with Spatial Smoothing (a special case of MSSL with a pre-defined precision matrix). Along with a lower error, MSSL captured spatial structure including long range teleconnections between some coastal cities.  The ability to harness this spatial structure and task relatedness within a GCM ensembles drives our attention toward MTL in other climate applications. 

Consider, in SD, each location in a region as a task with an identical set of possible covariates. These tasks are related through strong unknown spatial dependencies which can be harnessed for SD projections.  In the common high dimensional cases of SD, sparse features learned will provide greater significance as presented by~\cite{evgeniou2007multi}.  Furthermore, the structure between locations will be learned and may aid projections.  MSSL, presented by \cite{goncalves2014multi}, accounts for sparse feature selection and structure between tasks. 

In this study we aim to compare traditional statistical downscaling approaches, BCSD, Multiple Linear Regression, Lasso, and Support Vector Machines, against new approaches in machine learning, Multi-task Sparse Structure Learning and Convolutional Neural Networks (CNNs). During experimentation we apply common training architectures as part of the automated statistical downscaling framework. Results are then analyzed with a variety of metrics including, root mean Square error (RMSE), bias, skill of estimating underlying distributions, correlation, and extreme indices.

\section{Statistical Downscaling Methods}

\subsection{Bias Corrected Spatial Disaggregation}

BCSD~\cite{wood2002long} is widely used in the downscaling community due to its simplicity~\cite{abatzoglou2012comparison,Burger2012,wood2004hydrologic,maurer2010utility}. Most commonly, GCM data is bias corrected followed by spatial disaggregation on monthly data and then temporally disaggregated to daily projections. Temporal disaggregation is performed by selecting a month at random and adjusting the daily values to reproduce it's statistical distribution, ignoring daily GCM projections.  Thrasher et al. presented a process applying BCSD directly to daily projections~\cite{thrasher2012technical}, removing the step of temporal disaggregation. We the following steps with overlapping a reanalysis dataset and gridded observation data.  

	1) Bias correction of daily projections using observed precipitation. Observed precipitation is first remapped to match the reanalysis grid. For each day of the year values are pooled, $\pm$ 15 days, from the reanalysis and observed datasets to build a quantile mapping. With the quantile mapping computed, the reanalysis data points are mapped, bias corrected, to the same distribution as the observed data. When applying this method to daily precipitation detrending the data is not necessary because of the lack of trend and is therefore not applied. 

	2) Spatial disaggregation of the bias-corrected reanalysis data. Coarse resolution reanalysis is then bilinearly interpolated to the same grid as the observation dataset. To preserve spatial details of the fine-grained observations, the average precipitation of each day of the year is computed from the observation and set as scaling factors. These scaling factors are then multiplied to the daily interpolated GCM projections to provide downscaled GCM projections. 

\subsection{Automated Statistical Downscaling}

ASD is a general framework for statistical downscaling incorporating covariate selection and prediction~\cite{hessami2008automated}. Downscaling of precipitation using ASD requires two essential steps: 1. Classify rainy/non-rainy days ($\geq$ 1mm), 2. Predict precipitation totals for rainy days. The predicted precipitation can then be written as: 

\begin{equation}
\begin{aligned}
\label{eq:asd}
\textbf{E}[Y] = R * E[Y | R]
\text{ where } 
	R =
	    \begin{cases}
	      0, & \text{if}\ \textbf{P}(Rainy) < 0.5 \\
	      1, & \text{otherwise}
	    \end{cases}
\end{aligned}
\end{equation}
Formulating $R$ as a binary variable preserves rainy and non-rainy days. We test this framework using five pairs of classification and regression techniques.  

\subsubsection{Multiple Linear Regression}

The simplicity of Multiple Linear Regression (MLR) motivated its use in SD, particularly as part of SDSM~\cite{wilby2002sdsm} and ASD~\cite{hessami2008automated}. To provide a baseline relative to the following methods, we apply a variation of MLP using PCA. As discussed previously, PCA is implemented to reduce the dimensionality of a high dimensional feature space by selecting the components that account for a percentage (98\% in our implementation) of variance in the data. These principle components, $X$, are used as inputs to classify and predict precipitation totals. We apply a logistic regression model to classify rainy versus non-rainy days. MLP is then applied to rainy days to predict precipitation amounts, $Y$: 

\begin{equation}
\begin{aligned}
\label{eq:mlp}
& \hat{\beta} = \arg\!\min_{\beta}  \parallel Y - X\beta \parallel  \\
\end{aligned}
\end{equation}

This particular formulation will aid in comparison to ~\cite{Ghosh2010} where PCA is coupled with an SVM. 

\subsubsection{Elastic-Net}

Covariate selection can be done in a variety of methods, such as backward stepwise regression and partial correlations. Automatic covariate selection through the use of regularization terms, such as the $L_1/L_2$ norms in the statistical methods Lasso~\cite{tibshirani1996regression}, Ridge~\cite{hoerl1970ridge}, and Elastic-Net~\cite{zou2005regularization}.  Elastic-Net uses a linear combination of $L_1/L_2$ norms which we will apply in this intercomparison. Given a set of covariates $X$ and observations $Y$, Elastic-net is defined as:

\begin{equation}
\begin{aligned}
\label{eq:elnet}
& \hat{\beta} = \arg\!\min_{\beta} \big( \parallel Y - X\beta \parallel_2^2 + \lambda_1 \parallel \beta \parallel_1 +  \lambda_2 \parallel \beta \parallel_2^2 \big) \\
\end{aligned}
\end{equation}

The $L_1$ norm forces uninformative covariate coefficients to zero while the $L_2$ norm enforces smoothness while allowing correlated covariates to persist. Cross-validation is applied with a grid-search to find the optimal parameter values for $\lambda_1$ and $\lambda_2$. High-dimensional Elastic-Net is much less computational than stepwise regression techniques and most often leads to more generalizable models. A similar approach is applied to the classification step by using a logistic regression with an $L_1$ normalization term. Previous studies have considered the use of Lasso for SD~\cite{hammami2012predictor} but to our knowledge, none have considered Elastic-Net. 

\subsubsection{Support Vector Machine Regression}

Ghosh et al. introduced a coupled approach of PCA and Support Vector Machine Regression (SVR) for statistical downscaling~\cite{ghosh2008statistical,Ghosh2010}. The use of SVR for downscaling aims to capture non-linear effects in the data. As discussed previously, PCA is implemented to reduce the dimensionality of a high dimensional feature space by selecting the components that account for a percentage (98\% in our implementation) of variance in the data. Following dimensionality reduction, SVR is used to define the transfer function between the principle components and observed precipitation. Given a set of covariates (the chosen principle components) $X \in \mathbb{R}^{n \times m}$ and $Y \in \mathbb{R}^n$ with $d$ covariates and $n$ samples, the support vector regression is defined as~\cite{smola1997support}:

\begin{equation}
\begin{aligned}
& f(x) = \sum_{i=1}^d w_i \times K(x_i, x) + b
\end{aligned}
\end{equation}

where $K(x_i, x)$ and $w_i$ are the kernel functions and their corresponding weights with a bias term $b$. The support vectors are selected during training by optimizing the number of points from the training data to define the relationship between then predictand ($Y$) and predictors ($X$). Parameters $C$ and $\epsilon$ are set during training, which we set to $1.0$ and $0.1$ respectively, corresponding to regularization and loss sensitivity. A linear kernel function is applied to limit overfitting to the training set. Furthermore, support vector classifier was used for classification of rainy versus non-rainy days.

\subsubsection{Multi-task Sparse Structure Learning} 

Recent work in Multi-task Learning aims to exploit structure in the set of predictands while keeping a sparse feature set. Multi-task Sparse Structure Learning (MSSL) in particular learns the structure between predictands while enforcing sparse feature selection (\cite{goncalves2014multi}).  Goncalves et al. presented MSSL's exceptional ability to predict temperature through ensembles of GCMs while learning interesting teleconnections between locations (\cite{goncalves2014multi}). Moreover, the generalized framework of MSSL allows for implementation of classification and regression models. Applying MSSL to downscaling with least squares regression (logistic regression for classification), we denote $K$ as the number of tasks (observed locations), $n$ as the number of samples, and $d$ as the number of covariates with predictor $X \in \mathbb{R}^{n \times d}$, and predictand $Y \in \mathbb{R}^{n \times K}$.  As proposed in \cite{goncalves2014multi}, optimization over the precision matrix, $\Omegab$, is defined as

\begin{equation}
\label{eq:MSSL}
\begin{aligned}
\min_{\W,\Omegab \succ 0}  \bigg\{ \dfrac{1}{2} \sum_{k=1}^K \parallel X \W_k - Y_k \parallel_2^2 - \dfrac{K}{2} \text{log}|\Omegab| + Tr(\W \Omegab \W^T) + \lambda \parallel \Omegab \parallel_1 + \gamma \parallel \W \parallel_1   \bigg\}\\
\end{aligned}
\end{equation}

\noindent where $\W \in \mathbb{R}^{d \times K}$ is the weight matrix and $\Omegab \in \mathbb{R}^{K \times k}$ is an inverse precision matrix.  The $L_1$ regularization parameters $\lambda$ and $\gamma$ enforce sparsity over $\Omegab$ and $\W$. $\Omegab$ represents the structure contained between the high resolution observations.  Alternating minimization is applied to (\ref{eq:MSSL})


1. Initialize $\Omegab^0 = I_k, \W^0 = \textbf{0}_{d\Xb k}$ \\

2. for t=1,2,3,.. \textbf{do}

\begin{equation}
\W^{t+1} | \Omegab^t = \min_{\W}  \bigg\{ \dfrac{1}{2}\sum_{k=1}^K \parallel X_k \W_k -  Y_k \parallel_2^2 + Tr(\W \Omegab \W^T) + \gamma \parallel \W \parallel_1  \bigg\} \\
\label{eq:MSSL-W}
\end{equation}

\begin{equation}
\Omegab^{t+1} | \W^{t+1} = \min_{\Omegab} \bigg\{ Tr(\W \Omegab \W^T) - \dfrac{K}{2} log|\Omegab| + \lambda \parallel \Omegab \parallel_1 \bigg\} \\
\label{eq:MSSL-Omega}
\end{equation}

\noindent \ref{eq:MSSL-W} and \ref{eq:MSSL-Omega} are independently approximated through Alternating Direction Method of Multipliers (ADMM). Furthermore, by assuming the predictors of each task is identical (as it is for SD), \ref{eq:MSSL-W} is updated using Distributed-ADMM across the feature space \cite{boyd2011distributed}. 

MSSL enforces similarity between rows of $\W$ by learning the structure $\Omegab$.  For example, two locations which are nearby in space may tend to exhibit similar properties. MSSL will the exploit these properties and impose similarity in their corresponding linear weights. By enforcing similarity in linear weights, we are encouraging smoothness of SD projections between highly correlated locations.  $L_1$ regularization over $\W$ and $\Omegab$ jointly encourages sparseness and does not force structure. The parameters encouraging sparseness, $\gamma$ and $\lambda$, are chosen from a validation set using the grid-search technique. These steps are applied for both regression and classification. 

\subsubsection{Convolutional Neural Networks}

\begin{figure}
  \centering
  \includegraphics[width=\textwidth]{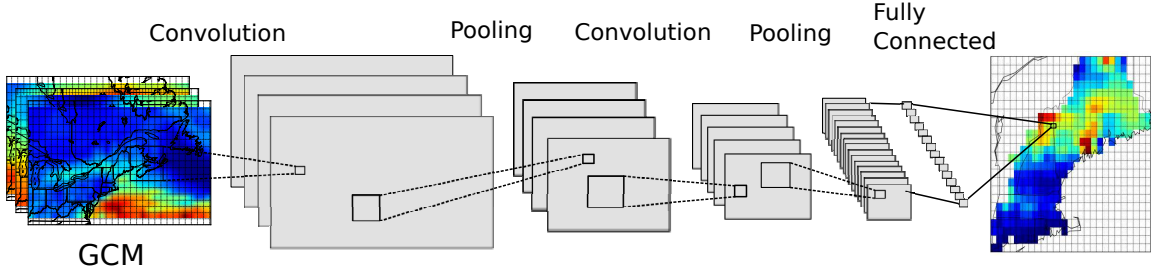}
  \caption{Given a set of GCM inputs $\Yb$, the first layer extracts a set of feature maps followed by a pooling layer. A second convolution layer is then applied to the reduced feature space and pooled one more time.  The second pooling layer is then flattened and fully connected to the high resolution observations.}
  \label{fig:cnn-framework}
\end{figure}

Artificial Neural Networks (ANN) have been widely applied to SD with mixed results~\cite{taylor2000quantile,schoof2001downscaling,Burger2012}, to name a few. In the past, ANNs had difficulty converging to a local minimum. Recent progress in deep learning has renewed interested in ANNs and are beginning to have impressive results in many applications, including image classification and speech recognition~\cite{krizhevsky2012imagenet,hinton2012deep,basu2015learning}.  In particular, Convolutional Neural Networks (CNNs) have greatly impacted computer vision applications by extracting, representing, and condensing spatial properties of the image~\cite{krizhevsky2012imagenet}.  SD may benefit from CNN advances by learning spatial representations of GCMs. Though CNNs rely on a high number of samples to reduce overfitting, dropout has been shown to be an effective method of reducing overfitting with limited samples~\cite{srivastava2014dropout}. We note that the number of observations available to daily statistical downscaling may cause overfitting.

CNNs rely on two types of layers, a convolution layer and a pooling layer. In the convolution layer, a patch of size $3 \times 3$ is chosen and slid with a stride of 1 around the image. A non-linear transformation is applied to each patch resulting in 8 filters. Patches of size $2 \times 2$ are then pooled by selecting the maximum unit with a stride of $2$.  A second convolution layer with a $3 \times 3$ patch to $2$ filters is followed by a max pooling layer of size $3 \times 3$ with stride $3$. The increase of pooling size decreases the dimensionality further.  The last pooling layer is then vectorized and densely connected to each high resolution location. This architecture is presented in Figure 1. 

Multiple variables and pressure levels from our reanalysis dataset are represented as channels in the CNN input.  Our CNN is trained using the traditional back propagation optimization with a decreasing learning rate. During training, dropout with probability 0.5 is applied the densely connected layer. This method aims to exploit the spatial structure contained in the GCM. A sigmoid function is applied to the output layer for classification. To our knowledge, this is the first application of CNNs to statistical downscaling. 

\subsection{Bias Corrected Spatial Disaggregation with MSSL}

To further understand the use of BCSD in Statistical Downscaling, we propose a technique to estimate the errors introduced in BCSD. As presented above, BCSD utilizes a relatively simple quantile mapping approach to statistical downscaling following by interpolation and spatial scaling. Following the BCSD estimates of the observed climate, we compute the presented errors, which may be consistent and have a predictive signal. Modeling such errors using the transfer function approaches above, such as MSSL, may uncover this signal and improve BCSD projections. To apply this technique, the following steps are taken:

\begin{enumerate}
\item Apply BCSD to the coarse scale climate variable and compute the errors.
\item Excluding a hold out dataset, use MSSL where they predictand is the computed errors and the predictands are from a different set of climate variables, such as Temperature, Wind, Sea Level Pressure, etc. 
\item Subtract the expected errors modeled by step 2 from BCSD projections in step 1. 
\end{enumerate}
\noindent The transfer function learned in step 2 is then applicable to future observations.

\section{Data}

The Northeastern United States endures highly variable season and annual weather patterns.  Variable climate and weather patterns combined with diverse topology provides difficulty in regional climate projection. Precipitation in particular varies heavily in frequency and intensity seasonally and annually~\cite{karl1998secular}. We choose this region to provide an in-depth comparison of statistical downscaling techniques for daily precipitation and extremes.  

\subsection{United States Unified Gauge-Based Analysis of Precipitation}

High resolution gridded precipitation datasets often provide high uncertainties due to a lack of gauge based observations, poor quality control, and interpolation procedures. Fortunately, precipitation gauge data in the continental United States is dense with high temporal resolution (hourly and daily).  The NOAA Climate Prediction Center CPC Unified Gauge-Based Analysis of Precipitation exploits the dense network of rain gauges to provide a quality controlled high resolution (0.25$^{\circ}$ by 0.25$^{\circ}$) gridded daily precipitation dataset from 1948 to the current date. State of the art quality control~\cite{chen2008assessing} and interpolation~\cite{xie2007gauge} techniques are applied giving us high confidence in the data. We select all locations within the northeastern United States watershed.

\subsection{NASA Modern-Era Retrospective Analysis for Research and Applications 2 (MERRA-2)}

Reanalysis datasets are often used as proxies to GCMs for statistical downscaling when comparing methods due to their low resolution gridded nature with a range of pressure levels and climate variables. Uncertainties and biases occur in each dataset, but state-of-the-art reanalysis datasets attempt to mitigate these issues. NASA's MERRA-2 reanalysis dataset~\cite{rienecker2011merra} was chosen after consideration of NCEP Reanalysis I/II~\cite{kalnay1996ncep} and ERA-Interm~\cite{dee2011era} datasets. \cite{kossin2015validating} showed the reduced bias of MERRA and ERA-Interm over NCEP Reanalysis II, which is most often used in SD studies.  MERRA-2 provides a significant temporal resolution from 1980 to present with relatively high spatial resolution (0.50$^{\circ}$ by 0.625$^{\circ}$). Satellite data provided by NASA’s GEOS-5 project in conjunction with NASA's data assimilation system when producing MERRA-2~\cite{rienecker2011merra}. 

Only variables available from the CCSM4 GCM model are selected as covariates for our SD models. Temperature, vertical wind, horizontal wind, and specific humidity are chosen from pressure levels 500hpa, 700hpa, and 850hpa.  At the surface level, temperature, sea level pressure, and specific humidity are chosen as covariates.  To most closely resemble CCSM4, each variable is spatially upscaled to 1.00$^{\circ}$ to 1.25$^{\circ}$ at a daily resolution. A large box centralized around the Northeastern Region ranging from 35$^{\circ}$ to 50$^{\circ}$ latitude and 270$^{\circ}$ to 310$^{\circ}$ longitude is used for each variable.  When applying the BCSD model, we use a spatially upscaled Land Precipitation MERRA-2 Reanalysis dataset at a daily temporal resolution.  Bilinear interpolation is applied over the coast to allow for quantile mapping of coastal locations as needed.  

\section{Experiments and Evaluation}


In-depth evaluation of downscaling techniques is crucial in testing and understanding their credibility. The implicit assumptions in SD must be clearly understood and tested when applicable. Firstly, SD models assume that the predictors chosen credibly represent the variability in the predictands. This assumption is partially validated through the choice of predictors presented above, which physically represents variability of precipitation. The remainder of the assumption must be tested through experimentation and statistical tests between downscaled projections and observations. The second assumption then requires the statistical attributes of predictands and predictors to be valid outside of the data using for statistical modeling. A hold out set will be used to test the feasibility of this assumption at daily, monthly, and annually temporal resolutions.  Third, the climate change signal must be incorporated in the predictands through GCMs. Predictands chosen for this experiment are available through CMIP5 CCSM4 simulations.  It is understood that precipitation is not well simulated by GCMs and therefore not used in ASD models~\cite{schiermeier2010real}. 

To test these assumptions, we provide in-depth experiments, analysis, and statistical metrics for each method presented above. The years 1980-2004 are used from training and years 2005-2014 are used for testing, taken from the overlapping time period of MERRA-2 and CPC Precipitation. For each method (excluding the special case of BCSD), we chose all covariates from each variable, pressure level, and grid point presented above, totaling 12,781 covariates. Each method applies either dimensionality reduction or regularization techniques to reduce complexity of this high dimensional dataset. Separate models are trained for each season (DJF, MAM, JJA, SON) and used to project the corresponding observations.

\noindent Analysis and evaluation of downscaled projections aim to cover three themes: 
\begin{enumerate}
\item Ability to capture daily anomalies.
\item Ability to respond to large scale climate trends on monthly and yearly temporal scales.
\item Ability to capture extreme precipitation events.  
\end{enumerate}

Similar evaluation techniques were applied in recent intercomparison studies of SD~\cite{Burger2012,gutmann2014intercomparison}. Evaluation of daily anomalies are tested through comparison of bias (Projected - Observed), Root Mean Square Error (RMSE), correlations, and a skill score~\cite{perkins2007evaluation}. The skill score presented by~\cite{perkins2007evaluation} measures how similar two probability density functions are from a range of 0 to 1 where 1 corresponds to identical distributions. Statistics are presented for winter (DJF), summer (JJA), and annually to understand season credibility. Statistics for spring and fall are computed but not presented in order to minimize overlapping climate states and simply results. Each of the measures are computed independently in space then averaged to a single metric. Large scale climate trends are tested by aggregating daily precipitation to monthly and annual temporal scales. The aggregated projections are then compared using Root Mean Square Error (RMSE), correlations, and a skill score as presented in~\cite{perkins2007evaluation}. Due to the limited number of data points in the monthly and yearly projections, we estimate each measure using the entire set of projections and observations. 

Climate indices are used for evaluation of SD models' ability to estimate extreme events. Four metrics from ClimDEX (http://www.clim-dex.org), chosen to encompass a range of extremes, will be utilized for evaluation, as presented by B\"{u}rger~\cite{Burger2012}.
\begin{enumerate}
\item CWD - Consecutive wet days $\geq$ 1mm
\item R20 - Very heavy wet days $\geq$ 20mm
\item RX5day - Monthly consecutive maximum 5 day precip
\item SDII - Daily intensity index = Annual total / precip days $\geq$ 1m
\end{enumerate}
Metrics will be computed on observations and downscaled estimates followed by annual (or monthly) comparisons. For example, correlating the maximum number of consecutive wet days per year between observations and downscaled estimates measures each SD models' ability to capture yearly anomalies. A skill score will also be utilized to understand abilities of reproducing statistical distributions. 

\section{Results}

Results presented below are evaluated using a hold-out set, years 2005-2014. Each model's ability to capture daily anomalies, long scale climate trends, and extreme events are presented. Our goal is to understand a SD model's overall ability to provide credible projections rather than one versus one comparisons, therefore statistical significance was not computed when comparing statistics.

\subsection{Daily Anomalies}

\begin{figure}
 \centerline{\includegraphics[width=\textwidth]{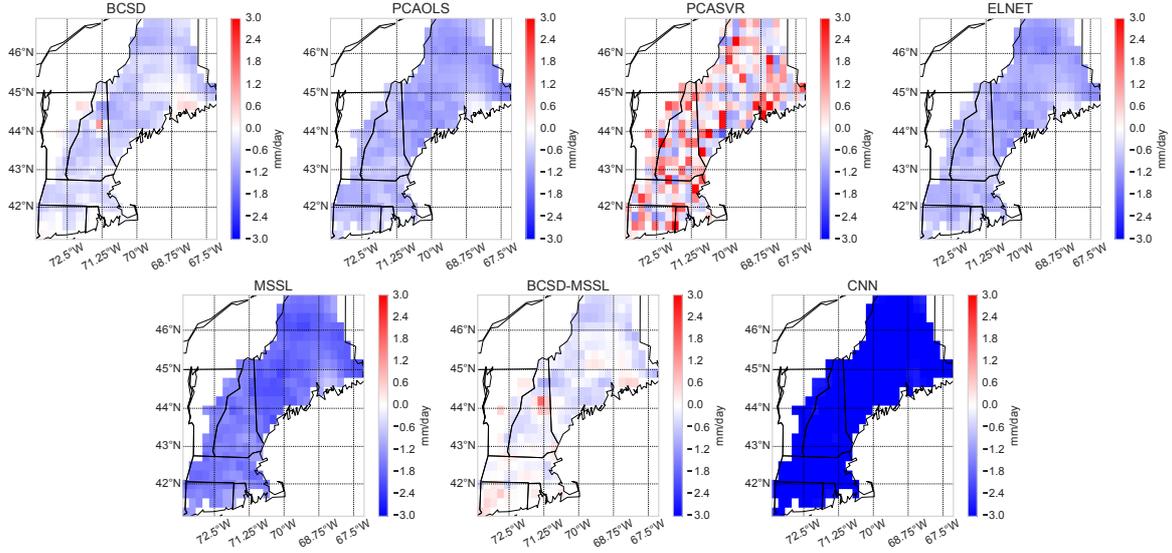}}
  \caption{Each map presents the spatial bias, or directional error, of the model. White represents no bias produced by the model while red and blue respectively show positive and negative biases.}
  \label{fig:bias}
\end{figure}

\begin{figure}
 \centerline{\includegraphics[width=\textwidth]{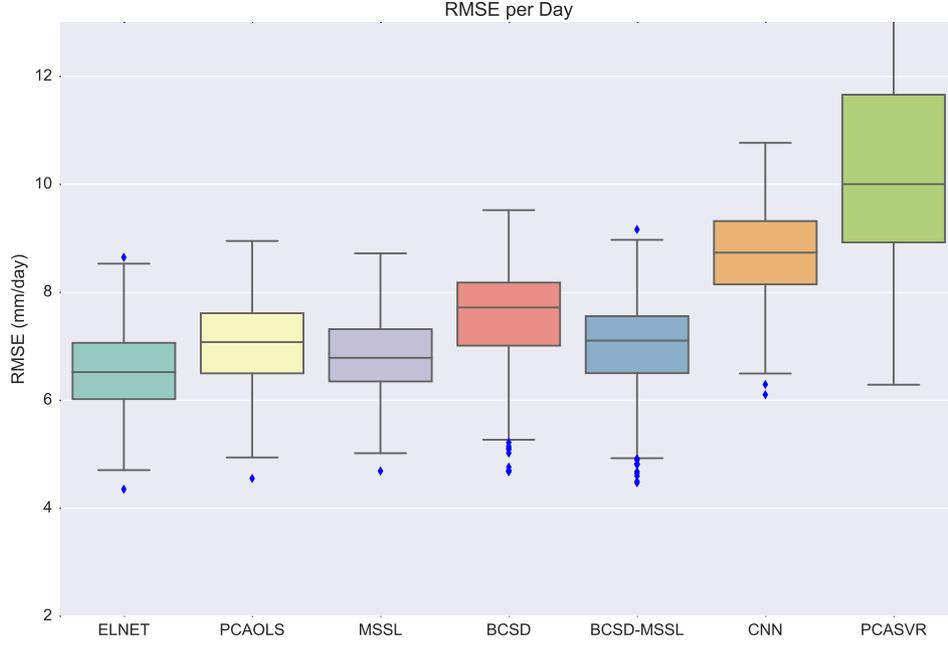}}
  \caption{Root mean square error (RMSE) is computed for each downscaling location and method. Each boxplot presents the distribution of all RMSEs for the respective method. The box shows the quartiles while the whiskers shows the remaining distribution, with outliers displayed by points.}
  \label{fig:daily-rmse}
\end{figure}

\begin{table}
\small
	\bigskip
	\noindent\makebox[\textwidth]{
	\begin{tabularx}{1.2\textwidth}{>{\em}l|XXX|XXX|XXX|XXX}
	\toprule
	{} &  \multicolumn{3}{c|}{Bias (mm)}    &  \multicolumn{3}{c|}{Correlation}   &     \multicolumn{3}{c|}{RMSE (mm)}   &   \multicolumn{3}{c}{Skill Score}      \\
	Season & Annual &    DJF &    JJA &      Annual &  DJF &  JJA & Annual &  DJF &  JJA & Annual &  DJF &  JJA \\
	\midrule
		BCSD      &  -0.44 & -0.36 & -0.36 &        0.52 &        0.49 &        0.46 &   0.75 & 0.65 & 0.81 &   0.93 &  0.92 &  0.89 \\
		PCAOLS    &  -0.89 & -0.71 & -1.16 &        0.55 &        0.60 &        0.49 &   0.70 & 0.55 & 0.75 &   0.82 &  0.81 &  0.76 \\
		PCASVR    &   0.37 &  0.04 &  0.20 &        0.33 &        0.39 &        0.31 &   1.10 & 0.79 & 1.05 &   0.91 &  0.87 &  0.87 \\
		ELNET     &  -0.88 & -0.66 & -1.16 &        0.64 &        0.69 &        0.55 &   0.65 & 0.50 & 0.72 &   0.84 &  0.85 &  0.78 \\
		MSSL      &  -1.58 & -1.20 & -2.05 &        0.62 &        0.64 &        0.54 &   0.68 & 0.55 & 0.74 &   0.92 &  0.90 &  0.88 \\
		BCSD-MSSL &  -0.16 & -0.10 & -0.02 &        0.58 &        0.60 &        0.50 &   0.69 & 0.56 & 0.77 &   0.79 &  0.80 &  0.74 \\
		CNN       &  -3.27 & -2.72 & -3.68 &        0.58 &        0.63 &        0.55 &   0.87 & 0.69 & 0.90 &   0.73 &  0.74 &  0.67 \\
	\bottomrule
	\end{tabularx}}
\caption{Daily statistical metrics averaged over space for annual, winter, and summer projections. Bias measures the directional error from each model. Correlation (larger is better) and RMSE (lower is better) describe the models ability to capture daily fluxuations in precipitation. The skill score statistic measure the model's ability to estimate the observed probability distribution.}
\label{tab:daily-stats}
\end{table}

Evaluation of daily anomalies depends on a model's ability to estimate daily precipitation given the state of the system. This is equivalent to analyzing the error between projections and observations.  Four statistical measures are used to evaluate these errors: bias, Pearson Correlation, skill score, and root mean square error (RMSE), as presented in Figure 2, Figure 3, and Table 1). All daily precipitation measures are computed independently in space and averaged to provide a single value. This approach is taken to summarize the measures as simply as possible.  Figure 2 shows the spatial representation of annual bias in Table 1. 

Overall, methods tend underestimate precipitation annually and seasonally with only PCASVR overestimating. BCSD-MSSL shows the lowest annual and summer bias and second lowest winter bias. BCSD is consistently under projects daily precipitation, but by modeling the possible error with MSSL, bias is reduced. PCAOLS and ELNET are less biased compared to MSSL. CNN has a strong tendency underestimate precipitation. Figure 2 shows consistent negative bias through space for BCSD, ELNET, PCAOLS, MSSL, and CNN while PCASVR shows no discernible pattern. 

Correlation measures in Table 1 presents a high linear relationship between projections and observations for the models ELNET (0.64 annually) and MSSL (0.62 annually).  We find that BCSD has a lower correlation even in the presence of error correction in BCSD-MSSL. PCASVR provides low correlations, averaging 0.33 annually, but PCAOLS performs substantially better at 0.55. 

The skill score is used to measure a model's ability to reproduce the underlying distribution of observed precipitation where a higher value is better between 0 and 1. BCSD, MSSL, and PCASVR have the largest skill scores, 0.93, 0.92, and 0.91 annually. We find that modeling the errors of BCSD decreases the ability to replicate the underlying distribution. The more basic linear models, PCAOLS and ELNET, present lower skill scores.  The much more complex CNN model has difficulty replicating the distribution.  

RMSE, presented in Figure 3 and Table 1, measures the overall ability of prediction by squaring the absolute errors.  The boxplot in Figure 3, where the box present the quartiles and whiskers the remaining distributions with outliers as points, shows the distribution of RMSE annually over space. The regularized models of ELNET and MSSL have similar error distributions and outperform others. CNN, similar to its under performance in bias, shows a poor ability to minimize error. The estimation of error produced by BCSD-MSSL aids in lowering the RMSE of plain BCSD. PCAOLS reasonably minimizes RMSE while PCASVR severely under-performs compared to all other models. Regression models applied minimize error during optimization while BCSD does not. Seasonally, winter is easier to project with summer being a bit more challenging.

\subsection{Large Climate Trends}

\begin{figure}
 \centerline{\includegraphics[width=\textwidth]{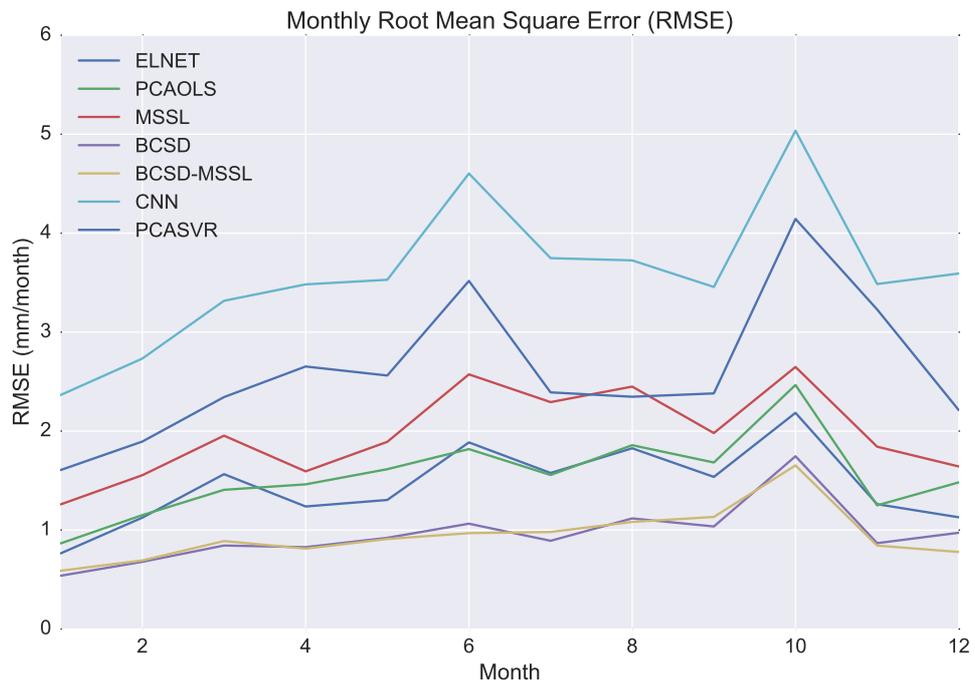}}
  \caption{The average root mean square Error for each month with each line representing a single downscaling model.}
  \label{fig:monthly-rmse}
\end{figure}

\begin{table}
\centering
\small
\bigskip
\begin{tabular}{>{\em}l|rr|rr|rr}
\toprule
{} &      \multicolumn{2}{c|}{RMSE (mm)}    &     \multicolumn{2}{c|}{Skill Score}   & \multicolumn{2}{c}{Correlation}        \\
Time-frame & Month & Year &  Month &    Year  &     Month & Year \\
\midrule
BCSD      &   31.97 &   204.78 &    0.88 &   0.63 &        0.85 &   0.64 \\
PCAOLS    &   50.01 &   362.73 &    0.75 &   0.27 &        0.63 &   0.41 \\
PCASVR    &   92.17 &   414.40 &    0.83 &   0.69 &        0.29 &   0.15 \\
ELNET     &   46.96 &   353.67 &    0.76 &   0.27 &        0.71 &   0.50 \\
MSSL      &   62.63 &   597.80 &    0.56 &   0.05 &        0.67 &   0.40 \\
BCSD-MSSL &   31.24 &   155.04 &    0.88 &   0.87 &        0.82 &   0.60 \\
CNN       &  112.21 & 1,204.27 &    0.01 &   0.00 &        0.59 &   0.54 \\
\bottomrule
\end{tabular}
\caption{Large Scale Projection Results: After aggregating daily downscaled estimates to monthly and yearly time scales, RMSE and Skill are computed per location and averaged.}
\label{tab:res-largscale}
\end{table}

Analysis of a SD model's ability to capture large scale climate trends can be done by aggregating daily precipitation to monthly and annual temporal scales.  To increase the confidence in our measures, presented in Table II and Figures 4 and 5, we compare all observations and projections in a single computation, rather than separating by location and averaging. 

Table 2 and Figure 4 show a wide range of RMSE. A clear difficulty in projecting precipitation in the fall, October in particular, is presented by each time-series in Figure 4. The difference in overall predictability relative to RMSE between the models is evident. BCSD and BCSD-MSSL have significantly lower monthly RMSEs compared to the others. Annually, BCSD-MSSL reduced RMSE by 25\% compared to plain BCSD. The linear models, ELNET, MSSL, and PCAOLS, have similar predictability while the non-linear models suffer, CNN being considerably worse. 

The skill scores in Table 2 show more difficulty in estimating the annual distribution versus monthly distribution. On a monthly scale BCSD and BCSD-MSSL skill scores outperform all other models but BCSD suffers slightly on an annual basis. However, BCSD-MSSL does not lose any ability to estimate the annual distribution. PCAOLS annual skill score is remarkably higher than the monthly skill score. Furthermore, the three linear models outperform BCSD on an annual basis. PCASVR's skill score suffers on an annual scale and CNN has no ability to estimate the underlying distribution. 

Correlation measures between the models and temporal scales show much of the same. BCSD has the highest correlations in both monthly (~0.85) and yearly (~0.64) scales while BCSD-MSSL are slightly lower. CNN correlations fall just behind BCSD and BCSD-MSSL. PCASVR fails with correlation values of 0.22 and 0.18.  ELNET has slightly higher correlations in relation to MSSL and PCAOLS. 

\subsection{Extreme Events}

\begin{figure}
 \centerline{\includegraphics[width=\textwidth]{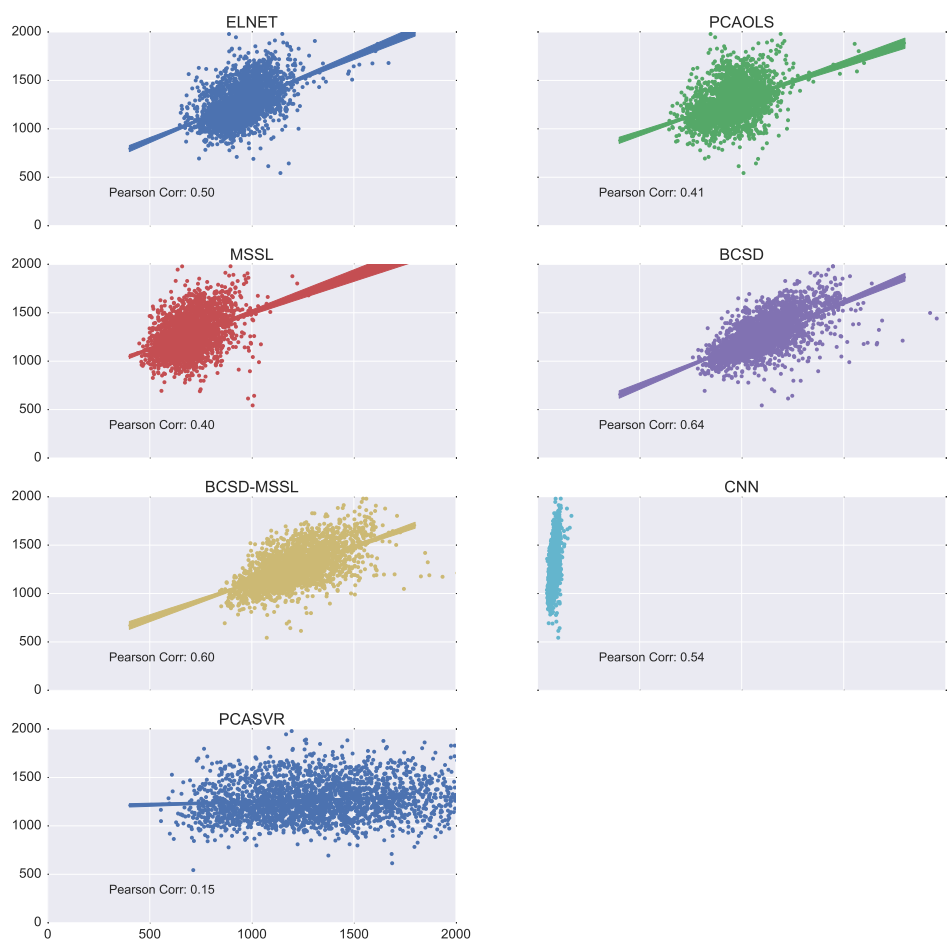}}
  \caption{Annual precipitation observed (x-axis) and projected (y-axis) for each model is presented along with the corresponding Pearson Correlation. Each point represents a single location and year.}
  \label{fig:annual-corr}
\end{figure}

\begin{figure}
 \centerline{\includegraphics[width=\textwidth]{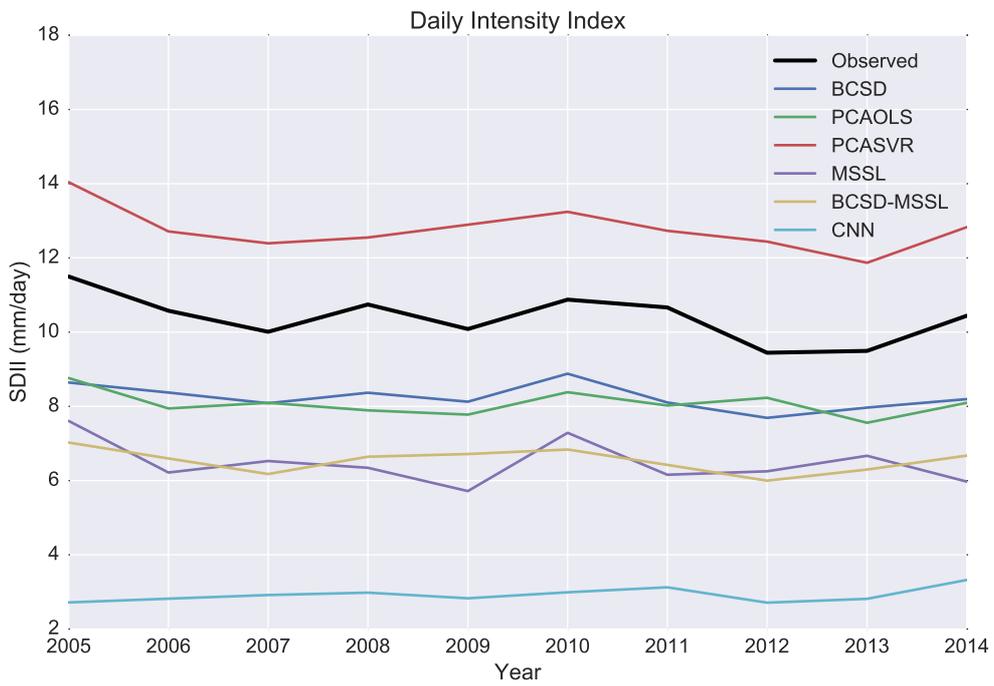}}
  \caption{The daily intensity index (Annual Precipitation/Number of Precipitation Days) averaged per year.}
  \label{fig:sdii}
\end{figure}

\begin{table}
	\centering
	\small
	\bigskip
	
	\begin{tabular}{>{\em}l|rrrr|rrrr}
	\toprule
		{} &  \multicolumn{4}{c|}{Correlation}  &  \multicolumn{4}{c}{Skill Score}     \\
		{Metric} &  CWD &  R20 & RX5day &  SDII &   CWD &  R20 & RX5day & SDII \\
	\midrule
		model     &      &      &        &       &       &      &        &      \\
		BCSD      & \textbf{0.43} & \textbf{0.83} &   \textbf{0.73} &  \textbf{0.70} &  0.71 & 0.80 &   \textbf{0.84} & 0.44 \\
		PCAOLS    & 0.25 & 0.65 &   0.44 &  0.67 &  0.69 & 0.60 &   0.65 & 0.44 \\
		PCASVR    & 0.24 & 0.81 &   0.19 &  0.25 &  0.78 & \textbf{0.89} &   0.80 & \textbf{0.65} \\
		ELNET     & 0.36 & 0.71 &   0.57 &  0.64 &  0.79 & 0.62 &   0.63 & 0.35 \\
		MSSL      & 0.33 & \textbf{0.84} &   0.56 &  0.52 &  \textbf{0.90} & 0.63 &   0.57 & 0.16 \\
		BCSD-MSSL & 0.25 & \textbf{0.83} &   0.70 &  \textbf{0.69} &  0.41 & 0.75 &   \textbf{0.84} & 0.08 \\
		CNN       & 0.07 &  --- &   0.33 & -0.30 &  0.05 & 0.59 &   0.01 & 0.00 \\
	\bottomrule
	\end{tabular}
	\caption{Statistics for ClimDEX Indices: For each model's downscaled estimate we compute four extreme indices, consecutive wet days (CWD), very heavy wet days (R20), maximum 5 day precipitation (RX5day), and daily intensity index (SDII), for each location.  We then compare these indices to those extracted from observations to compute correlation and skill metrics.}
	\label{tab:climdex}
\end{table}

A SD model's ability to downscale extremes from reanalysis depends on both the response to observed anomalies and ability to reproduce the underlying distribution. Resulting correlation measures present the response to observed anomalies, shown in Figure 6 and Table 3.  We find that BCSD has higher correlations for three metrics, namely consecutive wet days, very heavy wet days, and daily intensity index along with a similar results from 5-day maximum precipitation. Furthermore, modeling BCSD's expected errors with BCSD-MSSL decreases the ability to estimate the chosen extreme indices. Non-linear methods, PCASVR and CNN, suffer greatly in comparison to more basic bias correction and linear approaches. The linear methods, PCAOLS, ELNET, and MSSL, provide similar correlative performance. 

A skill score is used to quantify each method's ability to estimate an indices statistical distribution, presented in Table 3. Contrary to correlative results, PCASVR outperforms the other methods on two metrics, very heavy wet days and daily intensity index, with better than average scores on the other two metrics. BCSD also performs reasonably well in terms of skill scores while BCSD-MSSL suffers from the added complexity. MSSL estimates the number of consecutive wet days well but is less skilled on other metrics. The very complex CNN model has little ability to recover such distributions. 

Figure 6 displays a combination of correlative power and magnitude estimate of the daily intensity index. The SDII metric is computed from total annual precipitation and number of wet days.  A low SDII metric corresponds to either a relatively large number of estimated wet days or low annual precipitation. We find that the on average methods underestimate this intensity. Based on Figure 5 we see that CNN severely underestimates annual precipitation, causing a low SDII. In contrast, PCASVR overestimates annual precipitation and intensity.

Inconsistent results of PCASVR and CNN indicates that capturing non-linear relationships is outweighed by overfitting. However, BCSD and linear methods are more consistent throughout each metric.

\section{Discussion and Conclusion}

The ability of statistical downscaling methods to produce credible results is necessary for a multitude of applications. Despite numerous studies experimenting with a wide range of models for statistical downscaling, none have clearly outperformed others. In our study, we experiment with the off-the-shelf applicability of machine learning advances to statistical downscaling in comparison to traditional approaches. 

Multi-task Sparse Structure learning, an approach that exploits similarity between tasks, was expected to increase accuracy beyond automated statistical downscaling approaches. We find that MSSL does not provide improvements beyond ELNET, an ASD approach. Furthermore, the parameter set, estimated through cross-validation, attributed no structure aiding prediction. 

The recent popularity in deep learning along with it's ability to capture spatial information, namely Convolutional Neural Networks, motived us to experiment with basic architectures for statistical downscaling. CNNs benefit greatly by implicitly learning abstract non-linear spatial features based on the target variable. This approach proved to poorly estimate downscaled estimates relative to simpler methods. We hypothesize that implicitly learning abstract features rather than preserving the granular feature spaced caused poor performance. More experimentation with CNNs in a different architecture may still provide valuable results. 

BCSD, a popular approach to statistical downscaling, outperformed the more complex models in estimating underlying statistical distributions and climate extremes. In many cases, correcting BCSD's error with MSSL increased daily correlative performance but decreased skill of estimating the distribution. From this result, we can conclude that a signal aiding in prediction was lost during quantile mapping, interpolation, or spatial scaling. Future work may study and improve each step independently to increase overall performance. 

Of the seven statistical downscaling approaches studied, the traditional BCSD and ASD methods outperformed non-linear methods, namely Convolutional Neural Network and Support Vector regression, while downscaling daily precipitations.  We find that BCSD is skilled at estimating the statistical distribution of daily precipitation, generating better estimates of extreme events. The expectation of CNN and MSSL, two recent machine learning advances which we found most applicable to statistical downscaling, to outperform basic modeled proved false. Improvements and customization of machine learning methods is needed to provide more credible projections.

\section*{Acknowledgments}
This work was funded by NSF CISE Expeditions in Computing award 1029711, NSF CyberSEES award 1442728, and NSF BIGDATA award 1447587. 

MERRA-2 climate reanalysis datasets used were provided by the Global Modeling and Assimilation Office at NASA's Goddard Space Flight Center. The CPC Unified Gauge-Based Analysis was provided by NOAA Climate Prediction Center.

\bibliographystyle{plain}
\bibliography{references}

\begin{thebibliography}{10}

\bibitem{abatzoglou2012comparison}
John~T Abatzoglou and Timothy~J Brown.
\newblock A comparison of statistical downscaling methods suited for wildfire
  applications.
\newblock {\em International Journal of Climatology}, 32(5):772--780, 2012.

\bibitem{argyriou2007spectral}
Andreas Argyriou, Massimiliano Pontil, Yiming Ying, and Charles~A Micchelli.
\newblock A spectral regularization framework for multi-task structure
  learning.
\newblock In {\em Advances in neural information processing systems}, pages
  25--32, 2007.

\bibitem{basu2015learning}
Saikat Basu, Manohar Karki, Sangram Ganguly, Robert DiBiano, Supratik
  Mukhopadhyay, and Ramakrishna Nemani.
\newblock Learning sparse feature representations using probabilistic quadtrees
  and deep belief nets.
\newblock In {\em Proceedings of the European Symposium on Artificial Neural
  Networks, ESANN}, 2015.

\bibitem{boyd2011distributed}
Stephen Boyd, Neal Parikh, Eric Chu, Borja Peleato, and Jonathan Eckstein.
\newblock Distributed optimization and statistical learning via the alternating
  direction method of multipliers.
\newblock {\em Foundations and Trends{\textregistered} in Machine Learning},
  3(1):1--122, 2011.

\bibitem{Burger2012}
G.~B\"{u}rger, T.~Q. Murdock, a.~T. Werner, S.~R. Sobie, and a.~J. Cannon.
\newblock {Downscaling extremes-an intercomparison of multiple statistical
  methods for present climate}.
\newblock {\em Journal of Climate}, 25(12):4366--4388, June 2012.

\bibitem{chen2011integrating}
Jianhui Chen, Jiayu Zhou, and Jieping Ye.
\newblock Integrating low-rank and group-sparse structures for robust
  multi-task learning.
\newblock In {\em Proceedings of the 17th ACM SIGKDD international conference
  on Knowledge discovery and data mining}, pages 42--50. ACM, 2011.

\bibitem{chen2008assessing}
Mingyue Chen, Wei Shi, Pingping Xie, Viviane Silva, Vernon~E Kousky,
  R~Wayne~Higgins, and John~E Janowiak.
\newblock Assessing objective techniques for gauge-based analyses of global
  daily precipitation.
\newblock {\em Journal of Geophysical Research: Atmospheres}, 113(D4), 2008.

\bibitem{Coulibaly2005}
Paulin Coulibaly, Yonas~B. Dibike, and Fran\c{c}ois Anctil.
\newblock {Downscaling Precipitation and Temperature with Temporal Neural
  Networks}.
\newblock {\em Journal of Hydrometeorology}, 6(4):483--496, 2005.

\bibitem{dee2011era}
DP~Dee, SM~Uppala, AJ~Simmons, Paul Berrisford, P~Poli, S~Kobayashi, U~Andrae,
  MA~Balmaseda, G~Balsamo, P~Bauer, et~al.
\newblock The era-interim reanalysis: Configuration and performance of the data
  assimilation system.
\newblock {\em Quarterly Journal of the Royal Meteorological Society},
  137(656):553--597, 2011.

\bibitem{evgeniou2007multi}
A~Evgeniou and Massimiliano Pontil.
\newblock Multi-task feature learning.
\newblock {\em Advances in neural information processing systems}, 19:41, 2007.

\bibitem{frost2011comparison}
Andrew~J Frost, Stephen~P Charles, Bertrand Timbal, Francis~HS Chiew, Raj
  Mehrotra, Kim~C Nguyen, Richard~E Chandler, John~L McGregor, Guobin Fu,
  Dewi~GC Kirono, et~al.
\newblock A comparison of multi-site daily rainfall downscaling techniques
  under australian conditions.
\newblock {\em Journal of Hydrology}, 408(1):1--18, 2011.

\bibitem{frumkin2008climate}
Howard Frumkin, Jeremy Hess, George Luber, Josephine Malilay, and Michael
  McGeehin.
\newblock Climate change: the public health response.
\newblock {\em American Journal of Public Health}, 98(3):435--445, 2008.

\bibitem{ganguli2015water}
Poulomi Ganguli, Devashish Kumar, and Auroop~R Ganguly.
\newblock Water stress on us power production at decadal time horizons.
\newblock {\em arXiv preprint arXiv:1511.08449}, 2015.

\bibitem{Ghosh2010}
Subimal Ghosh.
\newblock Svm-pgsl coupled approach for statistical downscaling to predict
  rainfall from gcm output.
\newblock {\em Journal of Geophysical Research: Atmospheres}, 115(D22), 2010.

\bibitem{ghosh2008statistical}
Subimal Ghosh and PP~Mujumdar.
\newblock Statistical downscaling of gcm simulations to streamflow using
  relevance vector machine.
\newblock {\em Advances in water resources}, 31(1):132--146, 2008.

\bibitem{goncalves2014multi}
Andre~R Goncalves, Puja Das, Soumyadeep Chatterjee, Vidyashankar Sivakumar,
  Fernando~J Von~Zuben, and Arindam Banerjee.
\newblock Multi-task sparse structure learning.
\newblock In {\em Proceedings of the 23rd ACM International Conference on
  Conference on Information and Knowledge Management}, pages 451--460. ACM,
  2014.

\bibitem{gutmann2014intercomparison}
Ethan Gutmann, Tom Pruitt, Martyn~P Clark, Levi Brekke, Jeffrey~R Arnold,
  David~A Raff, and Roy~M Rasmussen.
\newblock An intercomparison of statistical downscaling methods used for water
  resource assessments in the united states.
\newblock {\em Water Resources Research}, 50(9):7167--7186, 2014.

\bibitem{haines2006climate}
Andy Haines, R~Sari Kovats, Diarmid Campbell-Lendrum, and Carlos Corval{\'a}n.
\newblock Climate change and human health: impacts, vulnerability and public
  health.
\newblock {\em Public health}, 120(7):585--596, 2006.

\bibitem{hammami2012predictor}
Dorra Hammami, Tae~Sam Lee, Taha~BMJ Ouarda, and Jonghyun Lee.
\newblock Predictor selection for downscaling gcm data with lasso.
\newblock {\em Journal of Geophysical Research: Atmospheres}, 117(D17), 2012.

\bibitem{hansen2013high}
Matthew~C Hansen, Peter~V Potapov, Rebecca Moore, Matt Hancher, SA~Turubanova,
  Alexandra Tyukavina, David Thau, SV~Stehman, SJ~Goetz, TR~Loveland, et~al.
\newblock High-resolution global maps of 21st-century forest cover change.
\newblock {\em science}, 342(6160):850--853, 2013.

\bibitem{hessami2008automated}
Masoud Hessami, Philippe Gachon, Taha~BMJ Ouarda, and Andr{\'e} St-Hilaire.
\newblock Automated regression-based statistical downscaling tool.
\newblock {\em Environmental Modelling \& Software}, 23(6):813--834, 2008.

\bibitem{hinton2012deep}
Geoffrey Hinton, Li~Deng, Dong Yu, George~E Dahl, Abdel-rahman Mohamed, Navdeep
  Jaitly, Andrew Senior, Vincent Vanhoucke, Patrick Nguyen, Tara~N Sainath,
  et~al.
\newblock Deep neural networks for acoustic modeling in speech recognition: The
  shared views of four research groups.
\newblock {\em Signal Processing Magazine, IEEE}, 29(6):82--97, 2012.

\bibitem{hoerl1970ridge}
Arthur~E Hoerl and Robert~W Kennard.
\newblock Ridge regression: Biased estimation for nonorthogonal problems.
\newblock {\em Technometrics}, 12(1):55--67, 1970.

\bibitem{jakob2011empirical}
Matthias Jakob~Theme{\ss}l, Andreas Gobiet, and Armin Leuprecht.
\newblock Empirical-statistical downscaling and error correction of daily
  precipitation from regional climate models.
\newblock {\em International Journal of Climatology}, 31(10):1530--1544, 2011.

\bibitem{kalnay1996ncep}
Eugenia Kalnay, Masao Kanamitsu, Robert Kistler, William Collins, Dennis
  Deaven, Lev Gandin, Mark Iredell, Suranjana Saha, Glenn White, John Woollen,
  et~al.
\newblock The ncep/ncar 40-year reanalysis project.
\newblock {\em Bulletin of the American meteorological Society},
  77(3):437--471, 1996.

\bibitem{karl1998secular}
Thomas~R Karl and Richard~W Knight.
\newblock Secular trends of precipitation amount, frequency, and intensity in
  the united states.
\newblock {\em Bulletin of the American Meteorological society},
  79(2):231--241, 1998.

\bibitem{kim2010tree}
Seyoung Kim and Eric~P Xing.
\newblock Tree-guided group lasso for multi-task regression with structured
  sparsity.
\newblock {\em International Conference on Machine Learning}, 2010.

\bibitem{kossin2015validating}
James~P Kossin.
\newblock Validating atmospheric reanalysis data using tropical cyclones as
  thermometers.
\newblock {\em Bulletin of the American Meteorological Society},
  96(7):1089--1096, 2015.

\bibitem{krizhevsky2012imagenet}
Alex Krizhevsky, Ilya Sutskever, and Geoffrey~E Hinton.
\newblock Imagenet classification with deep convolutional neural networks.
\newblock In {\em Advances in neural information processing systems}, pages
  1097--1105, 2012.

\bibitem{maurer2010utility}
Edwin~P Maurer, Hugo~G Hidalgo, T~Das, MD~Dettinger, and DR~Cayan.
\newblock The utility of daily large-scale climate data in the assessment of
  climate change impacts on daily streamflow in california.
\newblock {\em Hydrology and Earth System Sciences}, 14(6):1125--1138, 2010.

\bibitem{neumann2015climate}
James~E Neumann, Jason Price, Paul Chinowsky, Leonard Wright, Lindsay Ludwig,
  Richard Streeter, Russell Jones, Joel~B Smith, William Perkins, Lesley
  Jantarasami, et~al.
\newblock Climate change risks to us infrastructure: impacts on roads, bridges,
  coastal development, and urban drainage.
\newblock {\em Climatic Change}, 131(1):97--109, 2015.

\bibitem{parmesan2006ecological}
Camille Parmesan.
\newblock Ecological and evolutionary responses to recent climate change.
\newblock {\em Annual Review of Ecology, Evolution, and Systematics}, pages
  637--669, 2006.

\bibitem{perkins2007evaluation}
SE~Perkins, AJ~Pitman, NJ~Holbrook, and J~McAneney.
\newblock Evaluation of the ar4 climate models' simulated daily maximum
  temperature, minimum temperature, and precipitation over australia using
  probability density functions.
\newblock {\em Journal of climate}, 20(17):4356--4376, 2007.

\bibitem{rienecker2011merra}
Michele~M Rienecker, Max~J Suarez, Ronald Gelaro, Ricardo Todling, Julio
  Bacmeister, Emily Liu, Michael~G Bosilovich, Siegfried~D Schubert, Lawrence
  Takacs, Gi-Kong Kim, et~al.
\newblock Merra: Nasa's modern-era retrospective analysis for research and
  applications.
\newblock {\em Journal of Climate}, 24(14):3624--3648, 2011.

\bibitem{schiermeier2010real}
Quirin Schiermeier.
\newblock The real holes in climate science.
\newblock {\em Nature News}, 463(7279):284--287, 2010.

\bibitem{schoof2001downscaling}
Justin~T Schoof and SC~Pryor.
\newblock Downscaling temperature and precipitation: A comparison of
  regression-based methods and artificial neural networks.
\newblock {\em International Journal of Climatology}, 21(7):773--790, 2001.

\bibitem{smola1997support}
Alex Smola and Vladimir Vapnik.
\newblock Support vector regression machines.
\newblock {\em Advances in neural information processing systems}, 9:155--161,
  1997.

\bibitem{srivastava2014dropout}
Nitish Srivastava, Geoffrey Hinton, Alex Krizhevsky, Ilya Sutskever, and Ruslan
  Salakhutdinov.
\newblock Dropout: A simple way to prevent neural networks from overfitting.
\newblock {\em The Journal of Machine Learning Research}, 15(1):1929--1958,
  2014.

\bibitem{tatli2004statistical}
Hasan Tatli, H~N{\"u}zhet~Dalfes, and {\c{S}}~Sibel~Mente{\c{s}}.
\newblock A statistical downscaling method for monthly total precipitation over
  turkey.
\newblock {\em International Journal of Climatology}, 24(2):161--180, 2004.

\bibitem{taylor2000quantile}
James~W Taylor.
\newblock A quantile regression neural network approach to estimating the
  conditional density of multiperiod returns.
\newblock {\em Journal of Forecasting}, 19(4):299--311, 2000.

\bibitem{thrasher2012technical}
Bridget Thrasher, Edwin~P Maurer, C~McKellar, and PB~Duffy.
\newblock Technical note: Bias correcting climate model simulated daily
  temperature extremes with quantile mapping.
\newblock {\em Hydrology and Earth System Sciences}, 16(9):3309--3314, 2012.

\bibitem{tibshirani1996regression}
Robert Tibshirani.
\newblock Regression shrinkage and selection via the lasso.
\newblock {\em Journal of the Royal Statistical Society. Series B
  (Methodological)}, pages 267--288, 1996.

\bibitem{walther2002ecological}
Gian-Reto Walther, Eric Post, Peter Convey, Annette Menzel, Camille Parmesan,
  Trevor~JC Beebee, Jean-Marc Fromentin, Ove Hoegh-Guldberg, and Franz
  Bairlein.
\newblock Ecological responses to recent climate change.
\newblock {\em Nature}, 416(6879):389--395, 2002.

\bibitem{wilby2002sdsm}
Robert~L Wilby, Christian~W Dawson, and Elaine~M Barrow.
\newblock Sdsm—a decision support tool for the assessment of regional climate
  change impacts.
\newblock {\em Environmental Modelling \& Software}, 17(2):145--157, 2002.

\bibitem{wood2004hydrologic}
Andrew~W Wood, Lai~R Leung, V~Sridhar, and DP~Lettenmaier.
\newblock Hydrologic implications of dynamical and statistical approaches to
  downscaling climate model outputs.
\newblock {\em Climatic change}, 62(1-3):189--216, 2004.

\bibitem{wood2002long}
Andrew~W Wood, Edwin~P Maurer, Arun Kumar, and Dennis~P Lettenmaier.
\newblock Long-range experimental hydrologic forecasting for the eastern united
  states.
\newblock {\em Journal of Geophysical Research: Atmospheres}, 107(D20), 2002.

\bibitem{xie2007gauge}
Pingping Xie, Mingyue Chen, Song Yang, Akiyo Yatagai, Tadahiro Hayasaka,
  Yoshihiro Fukushima, and Changming Liu.
\newblock A gauge-based analysis of daily precipitation over east asia.
\newblock {\em Journal of Hydrometeorology}, 8(3):607--626, 2007.

\bibitem{zhang2012robust}
Tianzhu Zhang, Bernard Ghanem, Si~Liu, and Narendra Ahuja.
\newblock Robust visual tracking via multi-task sparse learning.
\newblock In {\em Computer Vision and Pattern Recognition (CVPR), 2012 IEEE
  Conference on}, pages 2042--2049. IEEE, 2012.

\bibitem{Zhang2015}
Xianliang Zhang and Xiaodong Yan.
\newblock A new statistical precipitation downscaling method with bayesian
  model averaging: a case study in china.
\newblock {\em Climate Dynamics}, 45(9-10):2541--2555, 2015.

\bibitem{zhang2012convex}
Yu~Zhang and Dit-Yan Yeung.
\newblock A convex formulation for learning task relationships in multi-task
  learning.
\newblock {\em Uncertainty in Artificial Intelligence}, 2012.

\bibitem{zou2005regularization}
Hui Zou and Trevor Hastie.
\newblock Regularization and variable selection via the elastic net.
\newblock {\em Journal of the Royal Statistical Society: Series B (Statistical
  Methodology)}, 67(2):301--320, 2005.

\end{thebibliography}

\end{document}